# Nine Features in a Random Forest to Learn Taxonomical Semantic Relations

**Enrico Santus*, Alessandro Lenci§, Tin-Shing Chiu*, Qin Lu*, Chu-Ren Huang***

\* The Hong Kong Polytechnic University, Hong Kong
esantus@gmail.com, cstschiu@comp.polyu.edu.hk, {qin.lu, churen.huang}@polyu.edu.hk
§ University of Pisa, Italy
alessandro.lenci@unipi.it

**Abstract**

ROOT9 is a supervised system for the classification of hypernyms, co-hyponyms and random words that is derived from the already introduced ROOT13 (Santus et al., 2016). It relies on a *Random Forest* algorithm and nine unsupervised corpus-based features. We evaluate it with a 10-fold cross validation on 9,600 pairs, equally distributed among the three classes and involving several Parts-Of-Speech (i.e. adjectives, nouns and verbs). When all the classes are present, ROOT9 achieves an *F1 score* of 90.7%, against a baseline of 57.2% (*vector cosine*). When the classification is binary, ROOT9 achieves the following results against the baseline: hypernyms-co-hyponyms 95.7% vs. 69.8%, hypernyms-random 91.8% vs. 64.1% and co-hyponyms-random 97.8% vs. 79.4%. In order to compare the performance with the state-of-the-art, we have also evaluated ROOT9 in subsets of the Weeds et al. (2014) datasets, proving that it is in fact competitive. Finally, we investigated whether the system learns the semantic relation or it simply learns the prototypical hypernyms, as claimed by Levy et al. (2015). The second possibility seems to be the most likely, even though ROOT9 can be trained on negative examples (i.e., switched hypernyms) to drastically reduce this bias.

## 1. Introduction

Distinguishing hypernyms from co-hyponyms and, in turn, discriminating them from semantically unrelated words (henceforth *randoms*) is a fundamental task in *Natural Language Processing* (NLP). Hypernymy in fact represents a key organization principle of semantic memory (Murphy, 2002), the backbone of taxonomies and ontologies, and one of the crucial semantic relations supporting lexical entailment (Geffet and Dagan, 2005). Co-hyponymy (or *coordination*) is instead the relation held by words sharing a close hypernym, which are therefore attributionally similar (Weeds et al., 2014).

The ability of discriminating hypernymy, co-hyponymy and random words has potentially infinite applications, including *Automatic Thesauri Creation*, *Paraphrasing*, *Textual Entailment*, *Sentiment Analysis* and so on (Weeds et al., 2014; Tungthamthiti et al. 2015). For this reason, in the last decades, numerous methods, datasets and shared tasks have been proposed to improve computers' ability in such discrimination, generally achieving promising results (Santus et al., 2016b; Roller et al., 2014, Weeds et al., 2014; Santus et al. 2014a; Rimmel, 2014; Lenci and Benotto, 2012; Kotlerman et al., 2010; Geffet and Dagan, 2005; Weeds and Weir, 2003). Both supervised and unsupervised approaches have been investigated. The former have been shown to outperform the latter in Weeds et al. (2014), even though Levy et al. (2015) have claimed that these methods may learn whether a term *y* is a prototypical hypernym, regardless of its actual relation with a term *x*.

In this paper we further investigate and revise ROOT13 (Santus et al., 2016b), a supervised method based on a *Random Forest* algorithm and thirteen corpus-based features. The feature contribution is evaluated with an ablation test, using a 10-fold cross validation on 9,600 pairs randomly extracted from *EVALution* (Santus et al., 2015) [1], *Lenci/Benotto* (Benotto, 2015) and *BLESS* (Baroni and Lenci, 2011). The ablation test has shown that four out of thirteen features were actually not contributing to the system's performance, and they were therefore removed, turning ROOT13 into ROOT9. On the 9,600 pairs, ROOT9 achieved an *F1 score* of 90.7% when the three classes were present, 95.7% when we had to discriminate hypernyms and co-hyponyms, 91.8% for hypernyms and randoms, and 97.8% for co-hyponyms and randoms.

In order to compare ROOT9 with the state-of-the-art, we have also evaluated it in the Weeds et al. (2014) datasets. Unfortunately, ROOT9 was not able to cover the full datasets, as several words in their pairs were missing from our *Distributional Semantic Model* (DSM) because of their low frequency. Nevertheless, the authors kindly provided the results of their models on our subsets, so that the comparison can be considered reliable. Also in relation to the state of the art, ROOT9 is proved to be competitive, being slightly outperformed in all the datasets only by the *svmCAT* model (Weeds et al., 2014), which is a *Support Vector Machine* (SVM) classifier run on the concatenation of the distributional vectors of the words in the pairs.

---

[1] The 9,600 pairs are available at https://github.com/esantus/ROOT9

Finally, we carried out an extra test to verify whether the system was actually learning the semantic relation between two word pairs, or simply identifying prototypical hypernyms (Levy et al. 2015). The test consisted in providing to the trained model switched hypernyms (e.g. from "*dog HYPER animal*" to "*dog RANDOM fruit*"), and verify how they were classified. Our results show that most of the switched hypernyms were in fact misclassified as hypernyms (especially when the words in the switched hypernyms were the same used to train the model on the real hypernyms), and that the only way to overcome such problem is to explicitly provide the model with bad examples (i.e., switched hypernyms tagged as randoms) during the training.

## 2. Related Work

Since the pioneering work of Hearst (1992), who used a pattern based approach for the "automatic acquisition of hyponyms from large text corpora", a large number of distributional methods were applied to the identification of hypernyms. These methods relied on interpretations of the *Distributional Hypothesis* (Harris, 1954), according to which the meaning of a linguistic expression can be inferred from its distribution in text corpora, so that linguistic expressions occurring in similar contexts are likely to be similar. These approaches, which can be either supervised or unsupervised, are generally implemented using *Vector Space Models* (VSMs; also called *Distributional Semantic Models*, DSMs), where vectors represent words, and their dimensions weight the association between the words and the contexts (Turney and Pantel, 2010).

Among the unsupervised methods, Weeds and Weir (2003) proposed the *Distributional Inclusion Hypothesis* (DIH), according to which less general words tend to occur in a subset of the contexts of their hypernyms. Their measure identified the direction of hypernymy with 71% accuracy on word-pairs extracted from WordNet (Fellbaum, 1998). This result, however, was not significantly better than the frequency baseline, according to which more general words are more frequent. Clarke (2009) extended the DIH, suggesting that generality difference can be calculated as the degree to which the narrower term's dimensions have lower values than the broader ones, across all the intersected dimensions. Lenci and Benotto (2012) adapted this measure to check not only to which extent the features of the narrower term are included in the features of the broader, but also how the features of the broader are not included in the features of the narrower. Kotlerman et al. (2010) combined *Average Precision* (AP) with the balancing approach of Szpektor and Dagan (2008), outperforming the above mentioned methods. Herbelot and Ganesalingam (2013) measured the Kullback-Leibler (KL) divergence between the probability distribution over context words for a term, and the background probability distribution, based on the idea that the smaller such KL was, the less informative the word was (and therefore more likely to be a hypernym). Rimmel (2014) considered the top features in a context vector as topics and used a *Topic Coherence* (TC) measure. Santus et al. (2014a) formulated the *Distributional Informativeness Hypothesis* (DInH), according to which the generality of a term can be inferred from the informativeness of its most typical linguistic contexts. In their evaluation, the authors have shown that hypernyms' most typical contexts are in fact less informative than hyponyms' ones.

Among the supervised methods, Baroni et al. (2012) proposed to use an SVM classifier on the concatenation (after having tried also subtraction and division) of the vectors. Roller et al. (2014) used the vectors' difference, while Weeds et al. (2014) implemented numerous combinations (difference, multiplication, sum, concatenation, etc.), comparing them against the most common unsupervised methods. The authors demonstrated that supervised methods generally perform better than unsupervised ones, but they acknowledge that these methods tend to learn ontological information, re-using it any time a word occur again in the dataset. For this reason, they suggest to adopt a new dataset, where words occur at most twice. Weeds et al. (2014)'s observation was further investigated by Levy et al. (2015), who claimed that supervised methods learn whether a term *y* is a prototypical hypernym, regardless of its actual relation with a term *x*.

## 3. Method

ROOT13 was firstly introduced in Santus et al. (2016b). It uses the *Random Forest* algorithm implemented in *Weka* (Breiman, 2001), with the default settings (i.e., 100 trees, 1 seed, and *maxDepth* and *numFeatures* initialized to 0), and relies on thirteen features that are carefully described below. Each of them is automatically extracted from a window-based DSM, trained on a combination of ukWaC and WaCkypedia corpora (about 2.7 billion words), counting word co-occurrences within the 5 nearest content words to the left and right of each target. Only adjectives, nouns and verbs with frequency above 1,000 are included in the DSM. As it will be shown in the evaluation, four out of thirteen features were redundant and were not contributing to the system performance. They were therefore dropped, turning ROOT13 into ROOT9.

### 3.1 Features

The feature set was designed to identify several distributional properties characterizing the terms in the pairs. On top of the standard distributional features (e.g., *co-occurrence frequency* and *words frequencies*), we have added several information that have been proved to be effective to discriminate paradigmatic semantic relations in vector spaces (Santus et al., 2014a; Santus et al., 2016a). All the features were computed using the above-mentioned DSM and normalized in the range 0-1.

### 3.1.1 Co-Occurrence

*Cooc* is defined as the co-occurrence frequency between the two terms in the pair, within the DSM window. According to the *Co-occurrence Hypothesis* (Charles and Miller, 1989), this measure is discriminative for synonyms and antonyms: antonyms are in fact expected to occur in the same sentence more often than synonyms. Since co-hyponyms can be often seen as a specific kind of opposition (e.g. "Winter or summer?"; Murphy 2003), this measure should help in discriminating them from hypernyms and randoms (Santus et al., 2014b-c).

### 3.1.2 Frequency

*Frequency* is an important property of words and it is a very discriminative information. For what concerns our task, Weeds and Weir (2003) have shown that the frequency baseline was very competitive in identifying the directionality of hypernymy-related pairs. We can therefore expect that hypernyms have higher frequency than hyponyms. Frequency is incorporated in our model with three features, namely one for each word involved in the pair (*Freq1,2*), plus one saving the difference between the frequencies (*Diff Freq*).

### 3.1.3 Entropy

*Entropy* is generally used to measure informativeness: the lower the entropy of an event, the higher its informativeness. Words occurrence in a corpus has very low entropy, as every word needs to fulfil certain morphological, syntactic and semantic requirements in order to occur in specific positions (e.g. in "$x$ barks", it is very likely that $x$ is "dog", because $x$ is expected to be a noun, and only dogs are known for barking). Nevertheless, words entropy varies according to several factors, such as the generality and prototypicality of the word. As claimed by Murphy (2002), the amount of informativeness in the taxonomies increases, when moving from the superordinate to the subordinate level. We use entropy as an index of word informativeness. It is calculated using the Shannon (1948)'s equation presented below:

$$H(w) = -\sum_{i=1}^{n} p(c_i|w) \cdot log_2(p(c_i|w))$$

where $p(c_i|w)$ is the probability of the context $c_i$ given the word $w$, computed as the ratio between the co-occurrence frequency of the pair $<w, c_i>$ and the total frequency of $w$.

In our system, entropy corresponds to three features, namely one for each word in the pair (*Entr1,2*), plus one saving the difference between the entropies (*Diff Entr*).

### 3.1.4 Shared and *APSyn*

*Shared* and *APSyn* (Santus et al., 2016a-b) are two features that do not rely on the full distribution of the words, but on the top $N$ most related contexts to the words in a pair, where $N$ was empirically fixed at 1000. The value of this parameter was tested in other experiments, some of which reported in Santus et al. (2016a).
Differently from Santus et al. (2016a-b), where the relation between contexts and words was measured with *Local Mutual Information* (LMI; Evert, 2005), in the current paper we used *Positive Pointwise Mutual Information* (PPMI; Levy et al., 2015), as it has shown some improvements:

$$PPMI(w,c) = \max(PMI(w,c), 0)$$

$$PMI(w,c) = \log_2\left(\frac{P(w,c)}{P(w) \times P(c)}\right) = \log_2\left(\frac{|w,c| \times D}{|w| \times |c|}\right)$$

were $w$ is the target word, $c$ is the given context, $P(w,c)$ is the probability of co-occurrence and $D$ is the collection of observed word-context pairs.
Once the PPMI values are assigned to all contexts of the target words (i.e. the words in the pair), we rank these contexts in a decreasing order, and consider only the top $N$, with $N = 1000$.
At this point, *Shared* is the cardinality of the intersection of the top $N$ contexts of the target words. *APSyn*, instead, is the weighted cardinality of the intersection, where the average ranking of the common features is used as a weight, as shown in the equation below:

$$APSyn(w_1, w_2) = \sum_{f \in N(F_1) \cap N(F_2)} \frac{1}{(rank_1(f) + rank_2(f))/2}$$

That is, for every feature *f* included in the intersection between the top $N$ features of $w_1$, $N(F_1)$, and $w_2$, $N(F_2)$, *APSyn* will add 1 divided by the average rank of the feature, among the top PPMI ranked features of $w_1$, $rank_1(f_1)$, and $w_2$, $rank_2(f_2)$.

### 3.1.5 Contexts Frequency

We have noticed that hypernyms tend to occur in more frequent contexts than co-hyponyms and randoms. Our system exploits two features, *C-Freq1,2*, capturing the frequency of the $N$ top contexts of the target words in the pair.

### 3.1.6 Contexts Entropy

Given what mentioned in 3.1.3 and the DIH and DInH (Weeds and Weir, 2003; Santus et al., 2014a), general words are likely to occur in a larger variety of contexts

(i.e. higher frequency) and in broader ones (i.e. less informative), compared to specific words. In fact, while hypernyms can certainly occur in narrower contexts, specific words are more likely to be chosen in these situations. Consider the following sentences:

a) The *X* has barked the all night.
b) The *Y* has arrested the thieves.

Any reader would agree that *X* is likely to be *dog* and *Y policeman*. Of course, *X* could have also been *animal* and *Y man*, or – even – both *X* and *Y* could have been *mammal*, but we expect that such general words are less frequently used in these contexts, as their hyponyms are more appropriate.

Adopting a similar approach to Santus et al. (2014a), we have measured the average entropy of the top *N* mostrelated contexts and used it as an index of generality. The higher the entropy, the less informative the word (i.e. it is more likely to be a hypernym). Our system uses one of these features for each target: *C-Entr1,2*.

## 4. Evaluation

### 4.1 Tasks

We have performed three tasks: i) an ablation test to evaluate the contribution of the features on our dataset (henceforth, *ROOT9 Dataset*; see Section 4.2); ii) an evaluation against the state of the art, and – in particular – against the best performant models in Weeds et al. (2014); iii) an evaluation on switched pairs to verify whether the actual semantic relations or the prototypical hypernyms (Levy et al., 2015) were learnt.

For what concerns the ablation test, we performed it on a tree-classes classification task (hypernyms, co-hyponyms and randoms), removing each feature at a time and measuring the loss/gain (*F1 score* is used for the evaluation on a 10-fold cross validation). Thanks to this task, we have found that four of our features were in fact redundant, and we have therefore removed them from the final model, turning ROOT13 into ROOT9. This is discussed in Section 5. Once the best model has been identified, we have performed three binary classification tasks, involving only two classes per time. *F1 score* on a 10-fold cross validation was chosen as accuracy measure. The second task, which is described in Section 6, consisted in binary classification tasks on the four datasets proposed by Weeds et al. (2014). These datasets are described below, in Section 4.3. The task allowed us to compare ROOT9 against the state of the art models reported in Weeds et al. (2014).

The last task is described in Section 7. It was performed on an extended *ROOT9 Dataset*, including also 3,200 randomly switched hypernyms to verify whether they were classified as hypernyms or as randoms.

### 4.2 *ROOT9* Dataset

We have used 9,600 pairs, randomly extracted from three datasets: *EVALution* (Santus et al., 2015), *Lenci/Benotto* (Benotto, 2015) and *BLESS* (Baroni and Lenci, 2011), which is freely available at https://github.com/esantus/ROOT9. The pairs are equally distributed among the three classes (i.e., hypernyms, co-hyponyms and random words) and involve several Parts-Of-Speech (i.e., adjectives, nouns and verbs).

The class of hypernyms contains 2,447 noun pairs, 458 verb pairs and 295 adjective pairs. The class of co-hyponyms has only 3,200 noun pairs, which were completely derived from BLESS, as this relation does not exist in the other two datasets. The class of randoms contains 1,100 noun pairs, 1,050 verb pairs and 1,050 random pairs.

The full dataset contains 4,263 terms (2,380 nouns, 958 verbs and 927 adjectives), so that every term occurs on average 4.5 times. Considering only the first word in the pairs, we have 1,265 different terms (987 nouns, 186 verbs and 92 adjectives). Considering instead only the second word, we have 3,665 terms (1,945 nouns, 860 verbs and 862 adjectives).

In the third task, we have extended this dataset randomly switching the 3,200 hypernymy pairs (e.g. from "*car HYPER vehicle*" to "*car RANDOM mammal*") to verify whether ROOT9 was able to classify them as randoms.

### 4.3 *Weeds* Dataset

In order to compare ROOT9 to the state-of-the-art, we have evaluated it with the datasets created by Weeds et al. (2014).[2] These are four datasets, containing respectively: i) hypernyms versus other relations (extracted from WordNet; henceforth *WN Hyper*); ii) co-hyponyms versus other relations (extracted from WordNet; henceforth *WN Co-Hyp*); iii) hypernyms versus other relations (extracted from BLESS; henceforth *Bless Hyper*); iv) co-hyponyms versus other relations (extracted from BLESS; henceforth *Bless Co-Hyp*).

The *WN* dataset (Weeds et al., 2014) – meaning both *WN Hyper* and *WN Co-Hyp* – in particular, was built after noticing that supervised systems tended to perform well also on random vectors. This happens because they are able to learn ontological information and re-use it whenever the words re-appear in other pairs. For this reason, the authors have constructed a dataset where words occurred at most twice (once on the left and once on the right of the relation). In this dataset, ontological information cannot be learnt and re-used, and indeed the random vectors cannot perform well.

Unfortunately our DSM did not cover the whole datasets, because of the chosen frequency threshold (in Table 1, we report the size of our subsets in comparison to the original datasets). However, Weeds et al. (2014) kindly provided

---
[2] The datasets are freely available at:
https://github.com/SussexCompSem/learninghypernyms

the results of their models on our subsets, so that the comparison is representative[3].

|  | WN Hyper | WN Co-Hyp | Bless Hyper | Bless Co-Hyp |
|---|---|---|---|---|
| *Weeds et al.* | *2514* | *4166* | *1668* | *5835* |
| Subset | 1791 | 2936 | 1636 | 5389 |
| Coverage % | 71.24 | 70.47 | 98.08 | 92.36 |

Table 1. Coverage on Weeds et al. (2014)'s datasets.

### 4.4 Baselines and Other Models

For our internal tests, we have implemented two baselines, which can be used as reference for evaluating the performance of ROOT9: *COSINE* and *RANDOM13*. The first baseline simply uses the *vector cosine* (*COSINE*) with a *Random Forest* classifier in the default settings (i.e. 100 trees, 1 seed, and *maxDepth* and *numFeatures* initialized to 0). This baseline is supposed to perform particularly well in discriminating similar words (i.e. hypernyms and co-hyponyms) from randoms. In fact, this measure has been extensively used to identify word similarity in vector spaces (Turney and Pantel, 2010) because it verifies the normalized correlation between the vectors of $w_1$ and $w_2$:

$$\cos(w_1, w_2) = \frac{\sum_{i=1}^n f_{1i} \times f_{2i}}{\sqrt{\sum (f_{1i})^2} \times \sqrt{\sum (f_{2i})^2}}$$

where $f_{xi}$ is the *i*-th dimension in the vector *x*.

The second baseline (*RANDOM13*) relies on a default *Random Forest* classifier, but uses thirteen randomly initialized features, with values between 0 and 1.

While the vector cosine achieves a reasonable accuracy, which is anyway far below the results obtained by our model, the random baseline performs much worst. The discrepancy with what found by Weeds et al. (2014) – namely that random vectors perform particularly well when words are re-used in the dataset – may depend on the small number of features, which does not allow the system to identify discriminative random dimensions.

In the second task (see Section 6), we have used as baselines the most competitive models reported in Weeds et al. (2014), namely the SVM classifiers trained on the PPMI vector of the second word (*svmSINGLE*), or on the concatenated (*svmCAT*), summed (*svmADD*), multiplied (*svmMULT*) and subtracted (*svmDIFF*) PPMI vectors of the words in the pair. Such vectors contain as features all major grammatical dependency relations involving open class Parts Of Speech. Also, the performance of three main unsupervised methods is reported as a reference: cosine (see above in this section), balAPinc (Kotlerman et al., 2010) and invCL (Lenci and Benotto, 2012). A threshold *p* empirically found in a training set was used in these methods for the decision,

---
[3] The subsets of Weeds et al. (2014)'s datasets are also available at https://github.com/esantus/ROOT9.

## 5. Task 1: Ablation Test and Binary Classification

Table 2 describes the features' contribution in the ablation test. Given the set of thirteen features of ROOT13 (Santus et al., 2016b), we have removed them one by one and measured the loss (negative) or the gain (positive).

|  | Hyper Co-Hyp Random | LOSS OR GAIN |
|---|---|---|
| **ROOT13** | **89.3** | 0.00% |
| - C-Freq 1, 2 | 88.2 | -1.23% |
| - C-Entr 1, 2 | 87.1 | -2.46% |
| *- APSyn* | *89.6* | *+0.34%* |
| *- Shared* | *89.6* | *+0.34%* |
| - Shared + APSyn | 87.7 | -1.79% |
| *- Diff Entr* | *89.6* | *+0.34%* |
| *- Diff Freq* | *89.7* | *+0.45%* |
| - Entr 1, 2 | 88.0 | -1.46% |
| - Freq 1, 2 | 88.3 | -1.12% |
| *- Cooc* | *89.4* | *+0.11%* |
| **ROOT9** | **90.7** | **+1.12%** |
| **BASELINES** | | |
| *ROOT9 using SMO* | *68.6* | *-23.18%* |
| *ROOT9 using Logistic* | *73.0* | *-18.25%* |
| COSINE | 57.2 | -35.95% |
| RANDOM13 | 33.4 | -62.60% |

Table 2. Ablation test, *F1 scores* on a 10-fold cross validation and loss/gain values. Scores are in percent.

As it can be easily seen from the table, most of features are contributing for an increment between 1.12% and 2.46%. The highest contribution comes from the *C-Entr1,2*, which were inspired at SLQS (Santus et al., 2014a), and the second highest contribute is given by *APSyn*, which was introduced in Santus et al. (2016a). Interestingly, four out of thirteen features were not contributing, penalizing the performance somewhere between 0.11% and 0.34%. These features are *Diff Entr*, *Diff Freq*, *Co-Occurrence*, and *APSyn* and *Shared*, when they are used together (so we kept only *APSyn*, removing *Shared*). The main reason why these features seem to affect negatively the results could be that they contain redundant information. If we remove both *APSyn* and *Shared*, for example, we have a loss of 1.79%, but when we remove only one of them we have a gain of 0.34%. In a similar way, *Diff Entr* and *Diff Freq* can be seen as redundant in respect to the features *Entr1,2* and *Freq1,2*. Perhaps surprisingly, *Cooc* does not contribute to the final score, and instead penalizes it.

Removing the four redundant features (we removed *Shared* but kept *APSyn*), ROOT13 turns into ROOT9. This system outperforms all the baselines (*COSINE*, *RANDOM13*) and ROOT13. For the sake of completeness, in Table 2 we also report the performance of ROOT9 using *Logistic Regression* (Cessie, 1992) and

*SMO* (Keerthi et al., 2001) classifiers. As it can be seen, the *Random Forest* version largely outperforms the other classifiers in this dataset. However, it is worth noticing here that such difference disappears with the *WN* datasets proposed by Weeds et al. (2014). See section 6, and – in particular – Table 4.

|  | Hyper Co-Hyp | Hyper Random | Co-Hyp Random |
|---|---|---|---|
| **ROOT13** | 94.3 | 91.1 | 97.4 |
| **ROOT9** | **95.7** | **91.8** | **97.8** |
| *- using SMO* | *77.3* | *80.1* | *93.0* |
| *- using Logistic* | *78.7* | *82.1* | *95.3* |
| COSINE | 69.8 | 64.1 | 79.4 |
| RANDOM13 | 50.1 | 49.6 | 51.4 |

Table 3. *F1 scores* on a 10-fold cross validation for binary classification tasks. Scores are in percent.

Table 3 describes the results of ROOT9 and the baseline in the binary classification tasks. These results confirm the analysis suggested above.

## 6. Task 2: ROOT9 vs. State of the Art

In Table 4, we show ROOT9's performance compared to the best systems reported by Weeds et al. (2014). The scores are all calculated on subsets of Weeds et al. (2014)'s datasets, as reported in Section 4.3.

Considering all the datasets, ROOT9 is the second best performing system, after *svmCAT* (Weeds et al., 2014), which uses the SVM classifier on the concatenation of PPMI vectors, containing as features all major grammatical dependency relations involving open class Parts Of Speech.

The SVM classifier on the sum (*svmADD*) and the multiplication (*svmMULT*) of the same PPMI vectors performs better in identifying co-hyponyms, but worst in identifying hypernyms. The SVM on the difference (*svmDIFF*) and on the second PPMI vector (*svmSINGLE*) is instead particularly good at identifying hypernyms, while it performs bad at identifying co-hyponyms.

Among the unsupervised methods, we report the results for the *cosine* and the methods of Lenci and Benotto (2012; *invCL*) and Kotlerman et al. (2010; *balAPinc*). Such methods classify using the best threshold *p* observed in the training sets. In general, unsupervised methods are less competitive.

Differently from what observed in Section 5, the performance of ROOT9 does not change by adopting a different classifier (i.e., *Random Forest*, *SMO* or *Logistic Regression*) on the *WN Hyper* and *WN Co-Hyp* datasets. However, it drastically changes again on the *BLESS Hyper* and *BLESS Co-Hyp* datasets. This may depend on the ability of the *Random Forest* classifier to learn more ontological information than *SMO* and *Logistic Regression*, also when the number of features is small.

|  | WN Hyper | WN Co-Hyp | Bless Hyper | Bless Co-Hyp |
|---|---|---|---|---|
| **ROOT9** | **69.8** | 60.8 | 94.6 | 87.7 |
| *- using SMO* | *67.7* | *60.9* | *65.5* | *70.4* |
| *- using Logistic* | *68.8* | *61.2* | *65.5* | *71.9* |
| **STATE OF THE ART (Weeds et al., 2014)** | | | | |
| **svmCAT** | **74.1** | **62.9** | **96.7** | **90.7** |
| **svmADD** | 40.9 | **66.0** | 68.5 | **94.1** |
| **svmMULT** | 40.3 | **63.2** | 75.1 | **96.4** |
| **svmDIFF** | **74.1** | 40.7 | 86.5 | 56.7 |
| **svmSINGLE** | 66.3 | 58.2 | **97.8** | 62.8 |
| cosine | 58.7 | 52.8 | 64.7 | 78.5 |
| **balAPinc** | 55.8 | 53.4 | 65.7 | 76.8 |
| **invCL** | 60.7 | **61.7** | 72.5 | 63.2 |

Table 4. *F1 scores*, in percent, on a 10-fold cross validation (state of the art models are evaluated on a 5-fold cross validation). {**bold**= best results vs. ROOT9; *italics* = other classifiers}.

## 7. Task 3: Learning Prototypical Hypernyms?

Finally, we have tried to test Levy et al. (2015)'s claim by evaluating the classifier on a dataset containing 3,200 hypernyms and 3,200 switched hypernyms (e.g. *apple* RANDOM *animal* and *dog* RANDOM *fruit*). In this evaluation, we have noticed that a large number of the switched hypernyms were indeed misclassified as hypernyms (up to 100% of them, if the words in the testing switched pairs were exactly the same used as hypernyms in the training set). In the attempt of correcting the behavior of the classifier, we extended the original 9,600 pairs dataset with other 3,200 switched hypernyms pairs labeled as randoms. It is relevant to notice that the switched hypernyms (tagged as *randoms*) contain the same words used in for the real hypernyms, and that in this new dataset, the size of the random class is double the others, including a total of 6,400 pairs. The new 10-fold cross validation test on the three classes registered a significant loss, passing from 90.7% to 84%. However, only 576 out of 6,400 randoms (most of which are likely to be the switched pairs) were misclassified as hypernyms.

## 8. Conclusions

In this paper, we have described ROOT9, a classifier for hypernyms, co-hyponyms and random words that is derived from an optimization of ROOT13 (Santus et al., 2016b). The classifier, based on the *Random Forest* algorithm, uses only nine unsupervised corpus-based features, which have been described, and their contribution assessed. The impressive results in our dataset, developed by randomly extracting 9,600 pairs from *EVALution* (Santus et al., 2015), *Lenci/Benotto* (Benotto, 2015) and *BLESS* (Baroni and Lenci, 2011), were further tested against the state-of-the-art models presented in Weeds et al. (2014). The comparison has shown that ROOT9 is in fact competitive with the state of the art, being outperformed on all the datasets only by an SVM trained on concatenated PPMI vectors. Interestingly, while on our dataset and on BLESS the chosen classifier is fundamental for the performance, on the *WN Hyper* and *WN Co-Hyp* datasets, *Random Forest*, *SMO* and *Logistic Regression* algorithm achieved a similar performance.

Finally, we have noticed the Levy et al. (2015)'s effect. However, we reduced it by training the model also on negative examples, namely switched hypernyms labeled as randoms (e.g. *apple* RANDOM *animal*, *dog* RANDOM *fruit*).

In future experiment, we plan to increase the number of features, investigating new distributional properties that may help in the classification without incurring in memorization effects such as those described by Levy et al. (2015).

## 9. Acknowledgements

We are very thankful to Julie Weeds for having helped us, recalculating the results of the Weeds et al. (2014) models also for our subsets of their datasets. Thanks also to Aristotelis Kostopoulos for the precious suggestions.

This work is partially supported by HK PhD Fellowship Scheme under PF12-13656

## 10. Main References